%% file: paper.tex
\documentclass{article} % For LaTeX2e
\usepackage{iclr2019_conference,times}
\pdfoutput=1

\input{math_commands.tex}

\usepackage{url}
\usepackage{comment}
\usepackage{subcaption}
\usepackage{graphicx}
\usepackage{layouts}
\usepackage{algorithm}
\usepackage{mathtools}
\usepackage{multirow}
\usepackage{todonotes}
\usepackage[noend]{algpseudocode}
\usepackage{soul}

\title{Generative predecessor models for sample-efficient imitation learning}

\author{Yannick Schroecker\thanks{This work was carried out at DeepMind.} \\
College of Computing\\
Georgia Institute of Technology\\
Atlanta, USA\\
\texttt{yannickschroecker@gatech.edu} \\
\And
Mel Vecerik \& Jonathan Scholz \\
DeepMind \\
London, United Kingdom\\
\texttt{\{vec,jscholz\}@google.com}
}
\iclrfinalcopy % Uncomment for camera-ready version, but NOT for submission.
\begin{document}

\maketitle
\begin{abstract}
We propose Generative Predecessor Models for Imitation Learning (GPRIL), a novel imitation learning algorithm that matches the state-action distribution to the distribution observed in expert demonstrations, using generative models to reason probabilistically about alternative histories of demonstrated states. We show that this approach allows an agent to learn robust policies using only a small number of expert demonstrations and self-supervised interactions with the environment. We derive this approach from first principles and compare it empirically to a state-of-the-art imitation learning method, showing that it outperforms or matches its performance on two simulated robot manipulation tasks and demonstrate significantly higher sample efficiency by applying the algorithm on a real robot.
    %\jsnote{how about: unlike SOTA approaches, has sample efficiency to run on real robot which we show...}
\end{abstract}
\section{Introduction}\label{sec:intro}

Training or programming agents to act intelligently in unstructured and sequential
environments is a difficult and central challenge in the field of artificial intelligence. Imitation
learning provides an avenue to tackle this challenge by allowing agents to learn from human teachers, which
constitutes a natural way for experts to describe the desired behavior and provides an efficient learning signal for the
agent. It is thus no surprise that imitation learning has enabled great
successes on robotic~\citep{chernova_robot_2014} as well as software domains (e.g.~\citet{aytar2018playing}).
Yet, key challenges in the field are diverse and include questions such as how to learn from observations
alone~(e.g.~\citet{aytar2018playing}), learning the correspondence between the expert's demonstrations 
and the agent's observations (e.g.~\citet{sermanet2017time}) as well as the question of how to integrate imitation
learning with other approaches such as reinforcement learning~(e.g.~\citet{vecerik2017leveraging}). 
However, at the core of the imitation learning problems lies the challenge of utilizing a given set of
demonstrations to match the expert's behavior as closely
as possible. 
In this paper, we approach this problem considering the setting where the set of expert demonstrations is given up-front and the dynamics of
the environment can only be observed through interaction.

In principle, imitation learning could be seen as a supervised learning problem, where the demonstrations
are used to learn a mapping from observed states to actions. This solution approach is known as behavioral cloning.
However, it has long been known that the sequential structure of the task admits more effective solutions. In
particular, the assumptions made in supervised
learning are restrictive and don't allow the agent to reason about the effect of its actions on it's future
inputs. As a result, errors and deviations from demonstrated behavior tend to accumulate over time as small mistakes lead the agent to parts of the observation
space that the expert has not explored~\citep{Ross2010}.
In this work, we propose a novel imitation learning algorithm, Generative Predecessor Models for Imitation Learning
(GPRIL), based on a simple core insight: 
Augmenting the training set with state-action pairs that %, under the current behavior, 
are likely to eventually lead the
agent to states demonstrated by the expert is an
effective way to train corrective behavior and to prevent accumulating errors.  

Recent advances in generative
modeling, such as~\citet{goodfellow2014generative, Kingma2013,Oord2016,van2016wavenet,
dinh2016density}, have shown great promise at modeling complex
distributions and can be used to reason probabilistically about such state-action pairs. Specifically, we propose to utilize Masked Autoregressive
Flows~\citep{papamakarios2017masked} to model long-term predecessor distributions, i.e. distributions over
state-action pairs which are conditioned on a state that the agent will see in the future. % and use such models to generate corrective, artificial training samples.
%%peng1993efficient edwards2018forward pan2018organizing goyal2018recall
Predecessor models have a long history in reinforcement learning (e.g.~\citet{peng1993efficient}) with recent approaches using deep networks to generate off-policy transitions~\citep{edwards2018forward,pan2018organizing} or to reinforce behavior leading to high-value states~\citep{goyal2018recall}. %In this work, we derive the use of such models from the perspective of state-action-distribution matching and propose a novel imitation learning algorithm that iterates over the following steps
Here, we use predecessor models to derive a principled approach to state-distribution matching and propose the following imitation learning loop:
%In this work, we  train such a predecessor model by utilizing self-supervised interactions with the
%environment and propose an algorithm that iterates over the following steps:
\begin{enumerate}
\item Interact with the environment and observe state, action as well as a future state. To encode long-term
corrective behavior, these states should be multiple steps apart.
\item Train a conditional generative model to produce samples like the observed state-action pair when conditioned on the observed
future state.
\item Train the agent in a supervised way, augmenting the training set using data drawn from the model 
conditioned on demonstrated states. The additional training data shows the agent how
to reach demonstrated states, enabling it to recover after deviating from expert behavior.
\end{enumerate}

In the above, we laid out the sketch of an algorithm that intuitively learns to reason about the states it will observe
in the future. In section \ref{sec:approach}, we derive this algorithm from first principles as
a maximum likelihood approach to matching the state-action distribution of the agent to the expert's
distribution. In section \ref{sec:eval}, we compare our approach %while providing a detailed account of our algorithm in section \ref{sec:approach}. 
to a state-of-the-art imitation learning method~\citep{ho2016b} and show that it matches or outperforms this baseline on our domains while being
significantly more sample efficient.
%Furthermore, we show that our approach can be used to learn effectively using demonstrated states alone which allows for a wider variety of methods that the expert can use to record demonstrations.
Furthermore, we show that GPRIL can learn using demonstrated states alone, allowing for a wider variety of methods to be used to record demonstrations.
Together these properties are sufficient to allow GPRIL to be applied in real-world settings, which we demonstrate in section \ref{sec:eval_robo}.  To our knowledge this is the first instance of dynamic, contact-rich and adaptive behavior being taught solely using the kinesthetic-teaching interface of a collaborative robot, without resorting to tele-operation, auxiliary reward signals, or manual task-decomposition. 

\section{Background}\label{sec:background}
\subsection{Markov decision processes without rewards}\label{sec:mdp}
As is usual, we model the problem as a Markov decision process without reward. That is, given state and action sets $\mathcal{S}, \mathcal{A}$, the agent is observing states $s
\in \mathcal{S}$ and taking actions $a \in \mathcal{A}$. 
In this work, we use $s$ and $a$ to refer to states and actions observed during
self-supervision and $\overline{s} \in \mathcal{S}$ and $\overline{a} \in \mathcal{A}$ to refer to target and
demonstration states and actions. We furthermore use superscripts $\overline{s}^{(i)}, \overline{a}^{(i)}$ to refer to specific instances, e.g. specific demonstrated state-action pairs, and subscripts, e.g. $s_t, a_t$, to indicate temporal sequences. 
The observed transitions are guided by the Markovian dynamics of the environment and the probability of transitioning
from state $s$ to state $s'$ by taking action $a$ is denoted as $p(s_{t+1}=s'|s_t=s, a_t=a)$.
The agent's behavior is defined by a stationary parametric policy $\pi_\theta(a|s)$ while the expert's behavior is
modeled by a stationary distribution $\pi^*(\overline{a}|\overline{s})$. We denote as $d_t^\pi(s)$ the probability of
observing state $s$ at time-step $t$ when following policy $\pi$. Under the usual ergodicity assumptions,
each such policy induces a unique stationary distribution of observed states $d^\pi(s)=\lim_{t\rightarrow \infty} d_t^\pi(s)$ as well as a stationary
joint state-action distribution ${\rho^\pi(s, a) := \pi(a|s)d^\pi(s)}$. Furthermore, we use $q_t^\pi$  to refer to the dynamics
of the time reversed Markov chain induced by a particular policy $\pi$ at time-step t 
\begin{equation}q_t^\pi(s_t=s,
    a_t=a|s_{t+1}=s') =
d_{t+1}(s')^{-1}d_t(s)\pi(a_t|s_t)p(s_{t+1}=s'|s_t=s, a_t=a)\end{equation}
and define $q^\pi(s_t=s,
a_t=a|s_{t+1}=s') := \lim_{t\rightarrow \infty}q_t^\pi(s_t=s, a_t=a|s_{t+1}=s')$. %For large $t \in \mathbb{N}$, we have $q_t^\pi(s_t=s, a_t=a|s_{t+1}=s') \approx q^\pi(s_t=s, a_t=a|s_{t+1}=s')$.
For the purposes of this work, we handle the episodic case with clear termination conditions by adding artificial transitions from terminal states to initial states. This creates a modified, ergodic MDP with identical state-distribution and allows us to assume arbitrarily large $t$ such that $q_t^\pi = q^\pi$. Finally, we extend this notation to multi-step transitions by writing $q^\pi(s_t=s, a_t=a|s_{t+j}=s')$.
\subsection{Imitation learning}\label{sec:imitation_learning}
% Definition of imitation learning. (Deduplicate with above?)
In this work, we are considering two settings of imitation learning. In the first setting, the agent is given a set of
observed states $\overline{s}^{(1)}, \overline{s}^{(2)}, \cdots, \overline{s}^{(N)}$ and observed corresponding actions 
$\overline{a}^{(1)}, \overline{a}^{(2)}, \cdots, \overline{a}^{(N)}$ as expert demonstrations. The goal in this setting is to learn a policy $\pi_\theta$ that matches the
expert's behavior as closely as possible. % Behavioral cloning and the problem of accumulating errors
In the second setting, the agent is given the states observed by the expert but is not aware of the actions the expert has taken. %, providing the human expert with more varied ways of recording demonstrations.
Recent years have seen heightened interest in a related setting where to goal is to track expert state trajectories~\citep{zhu2018reinforcement,peng2018deepmimic,pathak2018zero}. These approaches do not learn general policies that can adapt to unseen situations.
%A straightforward approach, usually referred to as behavioral cloning (BC),  trains a policy in the first setting by treating the task as a
%supervised learning problem (e.g. \citet{Pomerleau1989}). 
A straightforward approach to train general policies in the first setting, usually referred to as behavioral cloning, is to treat the task as a
supervised learning problem (e.g.~\citet{Pomerleau1989}).
However, as outlined in section \ref{sec:intro}, predictions
made by $\pi_\theta(a|s)$ influence future observations thus violating a key assumption of supervised learning, which states that inputs are drawn from an i.i.d. distribution. This has formally been analyzed by Ross
et al. who introduce a family of algorithms (e.g.~\citet{Ross2010, Ross2011a}) that provably avoid this issue. However,
these approaches require the expert to continuously provide demonstrations and thus are not applicable when the set of
demonstrations is assumed to be fixed.
% Solutions based on IRL
A popular avenue
        of research that considers this setting is inverse reinforcement learning (IRL). Inverse reinforcement
        learning~\citep{Ng2000} aims to
        learn a reward function for which $\pi^*$ is optimal and thus captures the intent of the expert. 
        %While most of these approaches are model-based, model free approaches exist, with the most popular approaches finding reward functions to match the state-action distribution to that of the demonstrations, breaking ties using the maximum entropy principle \cite{Boularias2011, Finn2016}
        The arguably most successful approach to IRL aims to match the state-action distribution to that of the demonstrations~\citep{Ziebart2007} with recent approaches extending these ideas to the model free case~\citep{Boularias2011,Finn2016,fu2018learning}.

        However, inverse reinforcement learning is indirect and ill-defined as many
        reward functions induce the same behavior. Recently, methods have been proposed that aim to
        match state-action distributions directly and achieve state-of-the-art
        result without learning a reward function first. Generative Adversarial Imitation Learning (GAIL)~\citep{ho2016b} uses an adversarial
        objective, training a discriminator to identify demonstrations and using TRPO to train a policy that fools the discriminator. While GAIL is able to achieve impressive results, the adversarial objective can make the learning procedure unstable and unpredictable. This
        is especially true when parts of the state-space are not under the agent's control, yielding a setting
        resembling a conditional GAN~\citep{mirza2014conditional}, which are prone to issues such mode collapse~\citep{odena2017conditional}. 
        %It should be noted that while GAIL aims to avoid the central problem of inverse reinforcement learning,
        %it has also inspired a new IRL method\cite{fu2018learning}. 
        We compare our approach with GAIL in section
        \ref{sec:eval}.
        State Aware Imitation Learning (SAIL)~\citep{Schroecker2017} is an alternative method that aims to learn the gradient of the state-action distribution using a temporal-difference
        update rule. This approach is able to avoid instabilities prompted by the adversarial learning rule but is only
        applicable to policies with a small number of parameters where learning a representation of the gradient is feasible. In this work, we follow
        a gradient descent approach similar to SAIL but estimate the gradient without representing it explicitly by a neural network.
        %\jsnote{The rest of this paragraph is confusing IMO, and doesn't help.}
        %These methods can also be used to learn from states alone by matching state distributions. While this does not
        %necessarily induce a unique policy,
        %including transition information such as velocities can allow the agent to learn solely from expert state trajectories nonetheless. 
        These methods can also be used to match state-distributions without actions. While the order of states is
        unspecified and the induced policy is therefore not unique, including transition information such as velocities
        can allow the agent to learn solely from expert state trajectories nonetheless.

        %The second setting considers the case where the agent is given the states observed by the expert but is not aware of the actions the expert has taken. This setting is more difficult but provides the human expert with more varied ways of recording demonstrations such as recording videos or kinesthetic teaching \cite{chernova_robot_2014}. 
        %Recent years have seen strong interest in this topic with a special emphasis on approaches that aim to track reference trajectories, using distance functions that are either task-specific\cite{} or learned \cite{}. However, in the general case where we aim to learn general policies that are able to adapt to unseen situations, distribution matching approaches such as IRL, GAIL, SAIL or GPRIL can be applied to match state-distributions as well.  While this does not necessarily induce a unique policy, including transition information, such as velocities, in the state-space can allow the agent to learn from expert state trajectories alone. 
        %These methods can also be applied to
        %the second setting by matching state distributions. While this does not necessarily induce a unique policy,
        %careful design of the state-space, e.g. including velocity information, can allow the agent to learn from expert state trajectories alone. 

\subsection{Generative models}\label{sec:cmaf}
% Overview of generative models
Recent years have seen great advances in deep generative models. A variety of approaches such as Generative Adversarial
    Networks~\citep{goodfellow2014generative}, Variational
    Auto Encoders~\citep{Kingma2013}, autoregressive networks (e.g.~\citet{germain2015made,Oord2016,van2016wavenet} and normalizing flows~\citep{dinh2016density} have been proposed which enable us to learn complex distributions and efficiently
    generate samples. 
   % These models extend readily to the case of modelling conditional distributions and have been used, for example, to model image distributions conditioned on a class label(e.g. \cite{mirza2014conditional,
   % odena2016conditional}).  
        In this work, we use generative models to model the distribution of long-term predecessor state-action pairs.
        While the approach we propose is model agnostic, we choose to model this distribution using masked
        autoregressive flows~\citep{papamakarios2017masked} (MAF). MAFs are trained using a maximum likelihood
        objective, which allows for
        a stable and straight-forward training procedure. 
        %However, approaches based on approximate objectives such as VAEs have shown promise on sequential domains as well (e.g. \cite{schmidhuber,tdvae}) and may also be used. 
        Autoregressive models are capable of representing complex distributions $p(x); x\in \mathbb{R}^n$ by factoring the distribution
        $p(x)=p_1(x_1) \prod_{i=1}^{N-1} p_{i+1}(x_{i+1}|x_1, \dots, x_{i})$ and learning a model for each $p_i$. In this paper,
        we model each $x_i$ to be distributed by $x_i\sim \mathcal{N}(\cdot|\mu_i, \sigma_i)$ where
        each $\mu_i$ and $\sigma_i$ is a function of $x_{1:i-1}$. 
        Masked autoencoders~\citep{germain2015made} provide a
        straight-forward approach to parameter sharing and allow representing these functions using a single
        network.  
        %While autoregressive models on their own can achieve impressive results~\cite{Ooord2016, van2016wavenet}, the order of
        %variables imposes a strong inductive bias on the model and can limit the expressiveness of the network.
        MAFs stack multiple autoregressive
        models with different orderings and thus avoid the strong inductive bias imposed by the order of variables. 
        % TODO explain this?
        Using the reparameterization trick, the autoregressive model can be seen as a deterministic
        and invertible transformation of a random variable: $x = f(z) := \mu +
        \sigma z; z\sim \mathcal{N}(\cdot|0, 1)$. The change of variable formula then allows us to calculate the
        density of $x$: \begin{equation}\label{eq:flows}\log p(x) = \log p_\mathcal{N}(f^{-1}(x)) + \log \det(|J(f^{-1}(x))|)\end{equation} Where the
        autoregressive nature of $f$ in eq. \ref{eq:flows} allows for tractable computation of the second term. 
        %This transformation can be applied to random variables distributed by arbitrary distributions with tractable density. 
        MAFs chain multiple such transformations to derive highly expressive explicit density models able to model complex dynamics between target states and long-term predecessor state-action pairs.
    \section{GPRIL}\label{sec:approach}
\begin{algorithm}[t]
\begin{algorithmic}[1]
    \small
    \Function{GPRIL}{$N_{\mathcal{B}}$, $N_{\pi}, B\;(\text{\it batch size})$}
    \For{$i \gets 0..\text{\#Iterations}$}
        \For{$k \gets 0..N_{\mathcal{B}}$}
            \For{$n \gets 0..B$}
                \State Sample $s^{(n)}_t, a^{(n)}_t$ from replay buffer
                \State Sample $j \sim Geom(1 - \gamma)$
                \State Sample $s^{(n)}_{t+j}$ from replay buffer
            \EndFor
            \State Update $\omega_s$ using gradient $\sum_{n=0}^{B}\nabla_{\omega_s} \log \mathcal{B}^s_{\omega_s}(s^{(n)}_t|s^{(n)}_{t+j})$
            \State Update $\omega_a$ using gradient $\sum_{n=0}^{B}\nabla_{\omega_a} \log \mathcal{B}^a_{\omega_a}(a^{(n)}_t|s^{(n)}_t, s^{(n)}_{t+j})$
        \EndFor

        \For{$k \gets 0..N_{\pi}$}
            \For{$n \gets 0..B$}
                \State Sample $\overline{s}^{(n)}, \overline{a}^{(n)}$ from expert demonstrations
                \State Sample $s^{(n)} \sim \mathcal{B}^s_{\omega_s}(\cdot|\overline{s}^{(n)})$, $a^{(n)} \sim \mathcal{B}^a_{\omega_a}(\cdot|s^{(n)}, \overline{s}^{(n)})$
            \EndFor
            \State Update $\theta$ using gradient $\sum_{n=0}^{B} \beta_\pi \nabla_{\theta} \log
            \pi_\theta(\overline{a}^{(n)}|\overline{s}^{(n)}) + \beta_d \nabla_{\theta} \log \pi_\theta(a^{(n)}|s^{(n)})$
        \EndFor

    \EndFor
    \EndFunction
\end{algorithmic}
\caption{Generative Predecessor Models for Imitation Learning (GPRIL)}
\label{fig:algorithm}
\end{algorithm}

In section \ref{sec:intro}, we provided an intuitive framework for using predecessor models to augment our
training set and achieve robust imitation learning from few samples.  In this section, we will derive this algorithm
based on state-action distribution matching. To this end, we first derive the gradient of the logarithmic state
distribution based on samples from a long-term predecessor distribution that we will define below. In section \ref{sec:model}, we
describe how to train a generative model of the predecessor distribution, which will allow us to evaluate this gradient.
Ascending on this gradient evaluated at demonstrated states leads
the agent to stick to those states and thus provides a corrective measure~\citep{Schroecker2017}. 
Furthermore, reproducing
the states of the expert can be sufficient to achieve the correct behavior if the state-space is chosen appropriately as
we will show in section \ref{sec:eval}. 
We will show how to use this gradient to match state-action-distributions in section \ref{sec:algorithm_overview}.

%TODO morimura using a derivation similar to \cite{TODOSAIL, TODOMorimura}. 
\subsection{Estimating the gradient of the state distribution}\label{sec:stationary_gradient}
Here, we will show that the samples drawn from a long-term predecessor distribution conditioned on
$\overline{s}$ enable us to estimate the gradient of the logarithmic state distribution $\nabla_\theta \log
d^{\pi_\theta}(\overline{s})$ and, later, to match the agent's state-action-distribution to that of the expert. 
To achieve this goal, we can utilize the fact that the stationary state distribution of a policy can be defined recursively in
terms of the state distribution at the previous time step, similar to~\citet{morimura2010derivatives}:
\begin{align}
    d^{\pi_\theta}(\overline{s}) = \int d^{\pi_\theta}(s) \pi_\theta(a|s) p(s_{t+1}=\overline{s}|s_t=s, a_t=a) ds,a.
\end{align}
Taking the derivative shows that this notion extends to the gradient
as well as its logarithm:
\begin{align}
    &\nabla_\theta d^{\pi_\theta}(\overline{s}) = \int \rho^{\pi_\theta}(s, a)
    p(s_{t+1}=\overline{s}|s_t=s, a_t=a)\left(\nabla_\theta \log d^{\pi_\theta}(s) + \nabla_\theta \log \pi_\theta(a|s)
    \right) ds,a\\
    &\nabla_\theta \log d^{\pi_\theta}(\overline{s}) = \int q^{\pi_\theta}(s_{t}=s, a_t=a|s_{t+1}=\overline{s})\left(\nabla_\theta \log d^{\pi_\theta}(s) + \nabla_\theta \log \pi_\theta(a|s) \right) ds,a \label{eq:prerollout}
\end{align}
The recursive nature of this gradient then allows us to unroll the gradient indefinitely. However, this process is cumbersome and will
be left for appendix \ref{sec:derivation}. We arrive at the following equality:
\begin{equation}\label{eq:path_expectation}
\nabla_\theta \log d^{\pi_\theta}(\overline{s}) = \lim_{T\rightarrow \infty} \int \sum_{j=0}^T  q^{\pi_\theta}(s_{t}=s, a_{t}=a| s_{t+j+1} = \overline{s}) \nabla_\theta \log
\pi_\theta(a|s) ds,a
\end{equation}
The derivation of our approach now rests on two key insights: 
\begin{comment}
First, while all past decisions contribute to the
probability of seeing state $\overline{s}$ at time step $t+j$, decisions that are made closer to time $t+j$ have a
significantly larger impact. In ergodic Markov chains such as the ones considered in this work, this intuition follows
from the behavior of Markov chains in the limit:
\begin{equation}\lim_{j\rightarrow \infty} \int q^{\pi_\theta}(s_t=s, a_t=a|s_{t+j}=\overline{s}) \nabla_\theta \log \pi_\theta (a|s) da,s =
\int d^{\pi_\theta}(s) \pi_\theta(a|s)\nabla_\theta \log \pi_\theta (a|s) da,s = 0.\end{equation} This result suggests that introducing a
discount factor that places higher weight on shorter sequences from $s$ to $\overline{s}$ can reduce variance while
introducing only a small bias.
Second, by introducing a discount factor, the effective time-horizon is now finite. This allows us to replace the
summation over infinite sequences as an expectation over state and actions sampled from these sequences.
Formally, we can write this as follows and arrive at our main result:
\end{comment}
First, in ergodic Markov chains, such as the ones considered in our setting, decisions that are made at time $t$
affect the probability of seeing state $\overline{s}$ at time $t+j$ more strongly if $j$ is small. In the limit, as $j \rightarrow
\infty$, the expectation of the gradient $\nabla_\theta \log \pi_\theta(a_t|s_t)$ vanishes and the decision at time $t$ only
adds variance to the gradient estimate. Introducing a discount factor $\gamma$ similar to common practice in
reinforcement learning~\citep{sutton1998reinforcement} places more emphasis on decisions that are closer in time and can thus greatly reduce variance. We explore this interpretation further in appendix \ref{sec:appendix_discount}.
Second, by introducing a discount factor, the effective time-horizon is now finite. This allows us to replace the
sum over all states and actions in each trajectory with a scaled expectation over state-action pairs. Formally, we can write this as follows and arrive at our main result:
\begin{equation}\label{eq:lsd} %TODO: index starting at 0 or 1, gamma starting at gamma^0 or gamma^1 in appendix
    \begin{aligned}
        \nabla_\theta \log d^{\pi_\theta}(\overline{s}) &\approx \int 
    \sum_{j=0}^\infty  \gamma^j q^{\pi_\theta}(s_{t}=s, a_{t}=a| s_{t+j+1} = \overline{s}) \nabla_\theta \log
\pi_\theta(a|s) ds,a \\
& \propto \mathbb{E}_{s, a\sim \mathcal{B}^{\pi_\theta}(\cdot, \cdot|\overline{s})}\left[\nabla_\theta \log \pi_\theta(a|s)\right]
    \end{aligned}
\end{equation}
where $\mathcal{B}^{\pi_\theta}$ corresponds to the long-term predecessor distribution modeling the distribution of states and
actions that, under the current policy $\pi_\theta$, will eventually lead to the given target state $\overline{s}$:
\begin{equation}\mathcal{B}^{\pi_\theta}(s,
    a|\overline{s}):=(1-\gamma)
    \sum_{j=0}^\infty  \gamma^j q^{\pi_\theta}(s_{t}=s, a_{t}=a| s_{t+j+1} = \overline{s})
\end{equation}

\subsection{Long-term generative predecessor models}\label{sec:model}
In the previous section, we derived the gradient of the logarithm of the stationary state distribution as approximately
proportional to the expected gradient of the log policy, evaluated at samples obtained from the long-term predecessor
distribution $\mathcal{B}^{\pi_\theta}$. In this work, we propose to train a model $\mathcal{B}_\omega^{\pi_\theta}$ to represent
$\mathcal{B}^{\pi_\theta}$ and use its samples to estimate $\nabla_\theta \log d^{\pi_\theta}(\overline{s})$. However, rather than unrolling a time-reversed Markov model in time, which is prone to accumulated errors, we propose to use a generative model to directly generate jumpy predictions. We have furthermore found that imposing a sensible order on autoregressive models achieves good results and thus propose to use two conditional MAFs~\citep{papamakarios2017masked} $\mathcal{B}_{\omega_s}^s, \mathcal{B}_{\omega_a}^a$ in a factored representation:% where $\mathcal{B}_{\omega_a}^a$ is the posterior, conditioning the policy on a state that is reached in the future: 
\begin{equation}\mathcal{B}_\omega^{\pi_\theta}(s, a|\overline{s}):=\mathcal{B}_{\omega_s}^s(s|\overline{s})\mathcal{B}_{\omega_a}^a(a|s,
\overline{s}).\end{equation} %TODO: differentiate between B and estimated B
To train this model, we collect training data using self-supervised roll-outs: We sample states, actions and target-states where the separation in time between the state and target-state is selected randomly based on the geometric distribution
parameterized by $\gamma$ as a training set for $\mathcal{B}_\omega^{\pi_\theta}$.

%\st{To describe this process in more detail, we use the current policy to obtain a sequence $s_0, a_0, s_1, a_1, \cdots$. In practice, we store this sequence in a short replay buffer which will be kept across multiple iterations.}
Training data for  $\mathcal{B}_{\omega_s}^s, \mathcal{B}_{\omega_a}^a$ are obtained by executing the current policy to obtain a sequence $s_0, a_0, s_1, a_1, \cdots$, which we store in a replay buffer.  
In practice, we store data from multiple iterations in this buffer in order to decrease the variance of the gradient.
While our algorithm does not explicitly account for off-policy samples, we found empirically that a short replay buffer does not degrade final performance while significantly improving sample efficiency.
To obtain a training sample, we first pick $s=s_t$ and $a=a_t$ for a random $t$. We now select %TODO: use s, a, overline{s} ?
a future state $\overline{s}=s_{t+j+1}$ from that sequence.
For any particular $s_{t+j+1}$ we now have $s, a \sim {q^{\pi_\theta}_t(s_t=\cdot, a_t=\cdot| s_{t+j+1}=\overline{s})}\approx
q^{\pi_\theta}(s_t=\cdot, a_t=\cdot| s_{t+j+1}=\overline{s})$. Note that in the episodic case, we can add transitions from terminal to initial
states and pick $t$ to be arbitrarily large such that the approximate equality becomes exact (as outlined in section \ref{sec:mdp}). In non-episodic
domains, we find the approximation error to be small for most $t$. Finally, we choose $j$ at random according to a
geometric distribution $j \sim Geom(1-\gamma)$ and have a training triple $s, a, \overline{s}$ that can be used to train $\mathcal{B}_{\omega_a}^a$ and $\mathcal{B}_{\omega_s}^s$ as it obeys
\begin{align}
    s, a \sim (1-\gamma) \sum_{j=0}^\infty \gamma^j q^{\pi_\theta}_t(s_t=\cdot, a_t=\cdot| s_{t+j+1}=\overline{s}) =
    \mathcal{B}^{\pi_\theta}(\cdot, \cdot| \overline{s}).
\end{align}
%Therefore, we use this sampling procedure to obtain training-data for $\mathcal{B}_{\omega_a}^a$ and $\mathcal{B}_{\omega_s}^s$.

\subsection{Matching state-action distributions with GPRIL} \label{sec:algorithm_overview}
State-action distribution matching has been a promising approach to sample-efficient and robust imitation learning
(see section \ref{sec:imitation_learning}). While each policy induces a unique distribution of states and behavioral cloning would therefore be sufficient in the limit, it is sub-optimal the case of limited data. Matching the joint-distribution directly ensures that we minimize discrepancies between
everything we observed from the expert and the behavior the agent exhibits. 
In this work, we propose a maximum likelihood based approach, % to the state-action distribution matching problem,
ascending on the estimated gradient of the joint distribution:
\begin{align}
    \nabla_\theta \log \rho^{\pi_\theta}(\overline{s}, \overline{a}) = \nabla_\theta
    \log \pi_\theta(\overline{a}|\overline{s}) + \nabla_\theta \log d^{\pi_\theta}(\overline{s}).
\end{align}
where $\nabla_\theta \log \pi_\theta(\overline{a}|\overline{s})$ can be computed directly by taking the gradient of the
policy using the demonstrated state-action pairs and $\nabla_\theta \log d^{\pi_\theta}(\overline{s})$ can be evaluated
using samples drawn from $\mathcal{B}_\omega^{\pi_\theta}(\cdot, \cdot| \overline{s})$ according to
equation \ref{eq:lsd}. We introduce scaling factors $\beta_\pi$ and $\beta_d$ to allow for finer control, interpolating between matching states only ($\beta_\pi=0$) and behavioral cloning ($\beta_d=0$) and have:
\begin{align}
    \nabla_\theta \log \rho^{\pi_\theta}(\overline{s}, \overline{a}) \approx \beta_\pi \log
    \pi_\theta(\overline{a}|\overline{s}) + \beta_d \mathbb{E}_{s, a\sim \mathcal{B}^{\pi_\theta}(\cdot, \cdot|\overline{s})}\left[\nabla_\theta \log \pi_\theta(a|s)\right].
\end{align}
Here,  higher values of $\beta_\pi$ provide more supervised guidance while lower values aim to prevent accumulating errors. This gives rise to the full-algorithm: We fill the replay buffer by asynchronously collecting experience using the current policy. Simultaneously, we repeatedly draw samples from the replay buffer to update
the predecessor models and use expert samples in combination with an equal number of artificial samples to update the
policy. This procedure is described fully in algorithm \ref{fig:algorithm}.

\section{Experiments}\label{sec:eval}
To evaluate our approach, we use a range of robotic insertion tasks similar to the domains introduced by~\citet{vecerik2017leveraging} but without access to a reward signal or, in some cases, expert actions. We choose these domains both for their practical use, and because they highlight challenges faced when applying imitation learning to the real world. 
Specifically, collecting experience using a robot arm is costly and demands efficient use of both demonstrations and autonomously gathered data.
Furthermore, insertion tasks typically require complex searching behavior and cannot be solved by open-loop tracking of a given demonstration trajectory when the socket position is variable.
%Insertion therefore poses a challenge for existing imitation learning and, especially, kinesthetic teaching approaches.
We first compare against state-of-the-art imitation learning methods on a simulated clip insertion task, then explore
the case of extremely sparse demonstrations on a simulated peg insertion task and finally, demonstrate real-world applicability on its physical counterpart.
\begin{figure}[t]
    \centering
    \begin{subfigure}[b]{0.38\textwidth}
        \center{\includegraphics[width=0.65\textwidth]{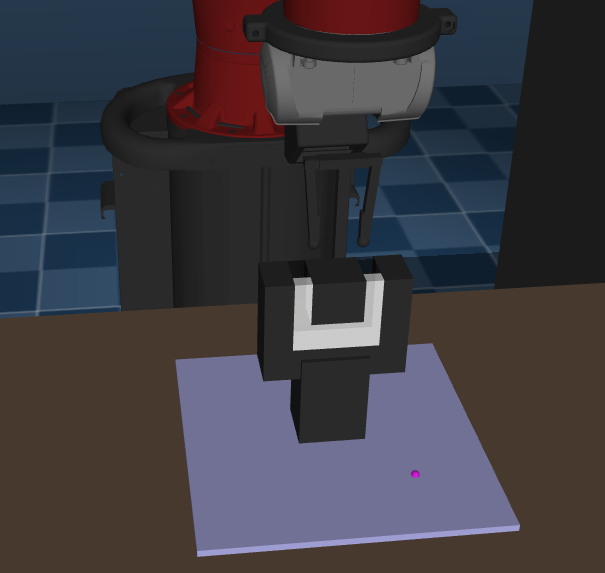}}
        \caption{}
        \label{fig:clippy_robo}
    \end{subfigure}
    \begin{subfigure}[b]{0.480\textwidth}
        \center{\includegraphics[width=0.95\textwidth]{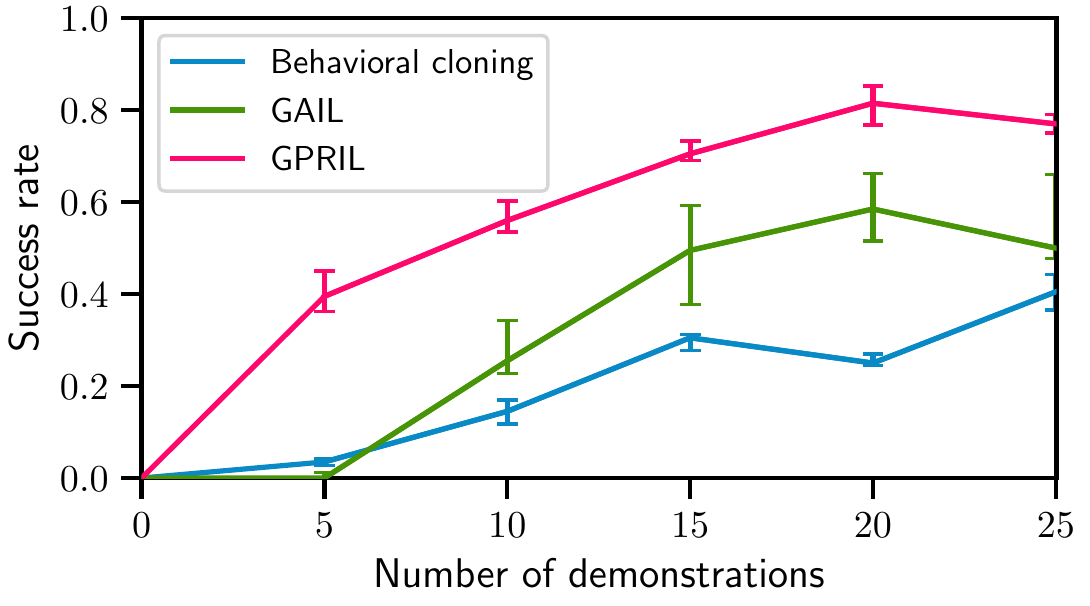}}\vspace*{-0.25cm}
        \caption{}
        \label{fig:clippy_numdemo}
    \end{subfigure}
%\caption{\textbf{a)} Depiction of the clip-insertion task. \textbf{b)} Median final success
%rate and interquartile range out of 100 roll-outs over 8 seeds. GPRIL achieves the highest success rate %followed by GAIL.}
%\end{figure}
%\begin{figure}[t]
    \begin{subfigure}[b]{0.480\textwidth}
        \center{\includegraphics[width=\textwidth, height=0.55\textwidth]{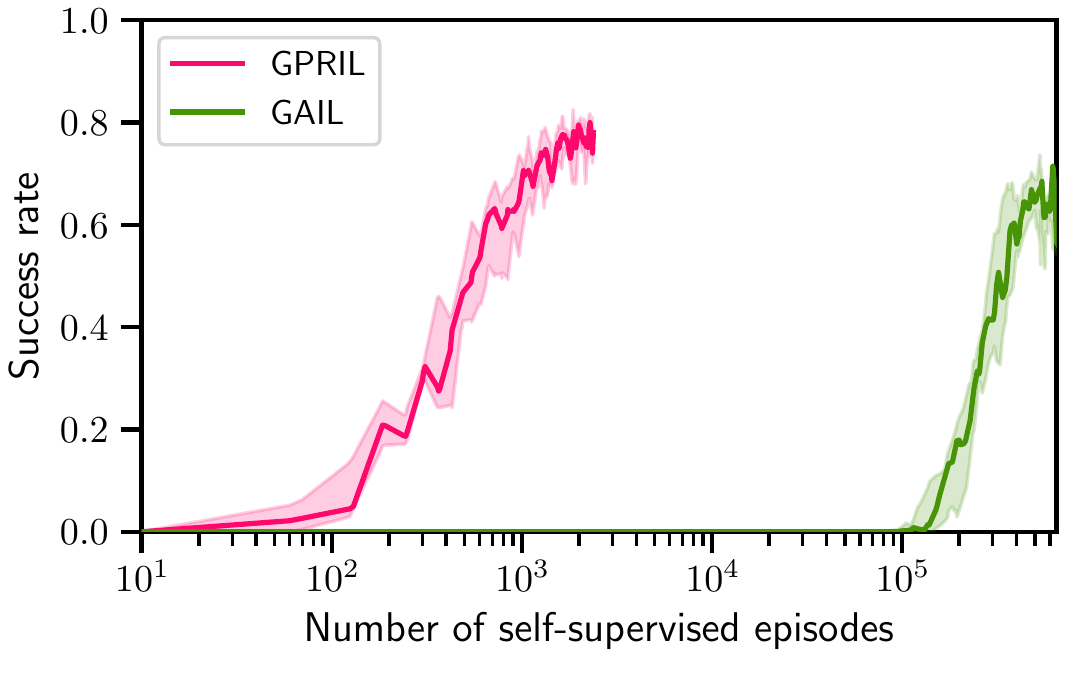}}%\the\textwidth %196.75311 pt = 0.495\textwidth %\printinunitsof{in}\prntlen{\textwidth} 2.723in
        \vspace*{-0.3cm}\caption{}\vspace*{-0.1cm}
        \label{fig:clippy_samples}
    \end{subfigure}
    \begin{subfigure}[b]{0.480\textwidth}
        \center{\includegraphics[width=\textwidth, height=0.55\textwidth]{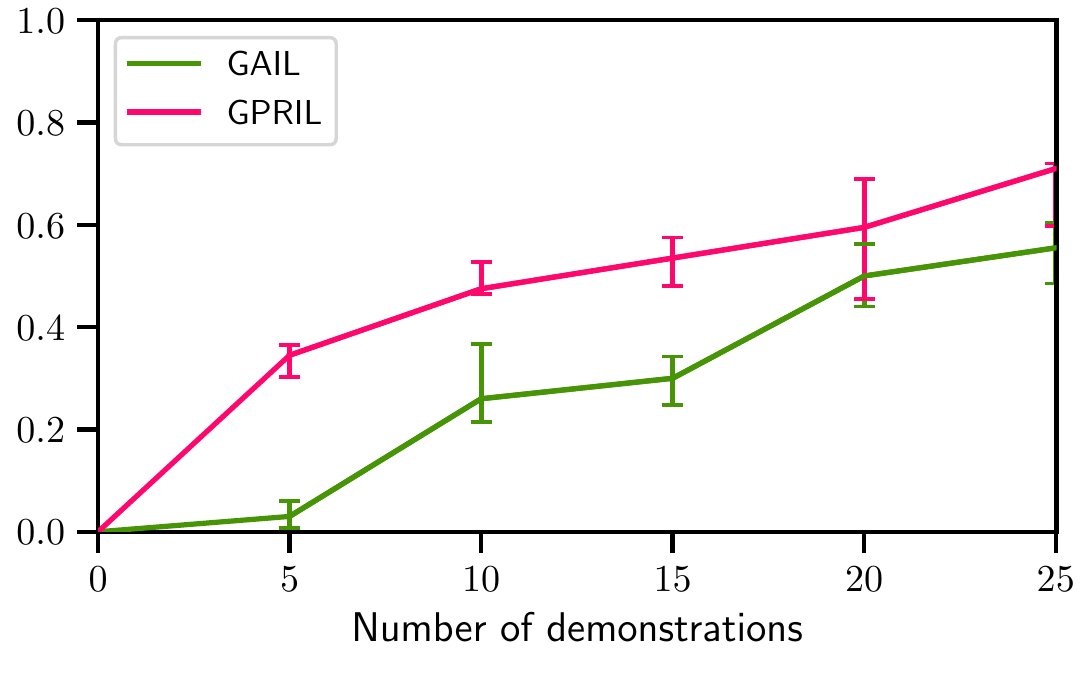}}\vspace*{-0.3cm}
        \caption{}\vspace*{-0.1cm}
        \label{fig:clippy_obs}
    \end{subfigure}
    \caption{\textbf{a)} Depiction of the clip-insertion task. \textbf{b)} Median final success
rate and interquartile range out of 100 roll-outs over 8 seeds. GPRIL achieves the highest success rate followed by GAIL. \textbf{c)} Median final success rate and IQR on clip insertion comparing sample efficiency. GPRIL is able
    to solve the task using several orders of magnitude fewer environment interactions. \textbf{d)} Comparison on clip
insertion trained on states alone. Learning from states alone only slightly affects performance.}
\end{figure}
\begin{figure}[t]
    \begin{subfigure}[b]{0.3\textwidth}
        %\center{\includegraphics[width=\textwidth]{figures/gears_robo.png}\vspace*{0.7cm}}
        \center{\includegraphics[width=0.463\textwidth]{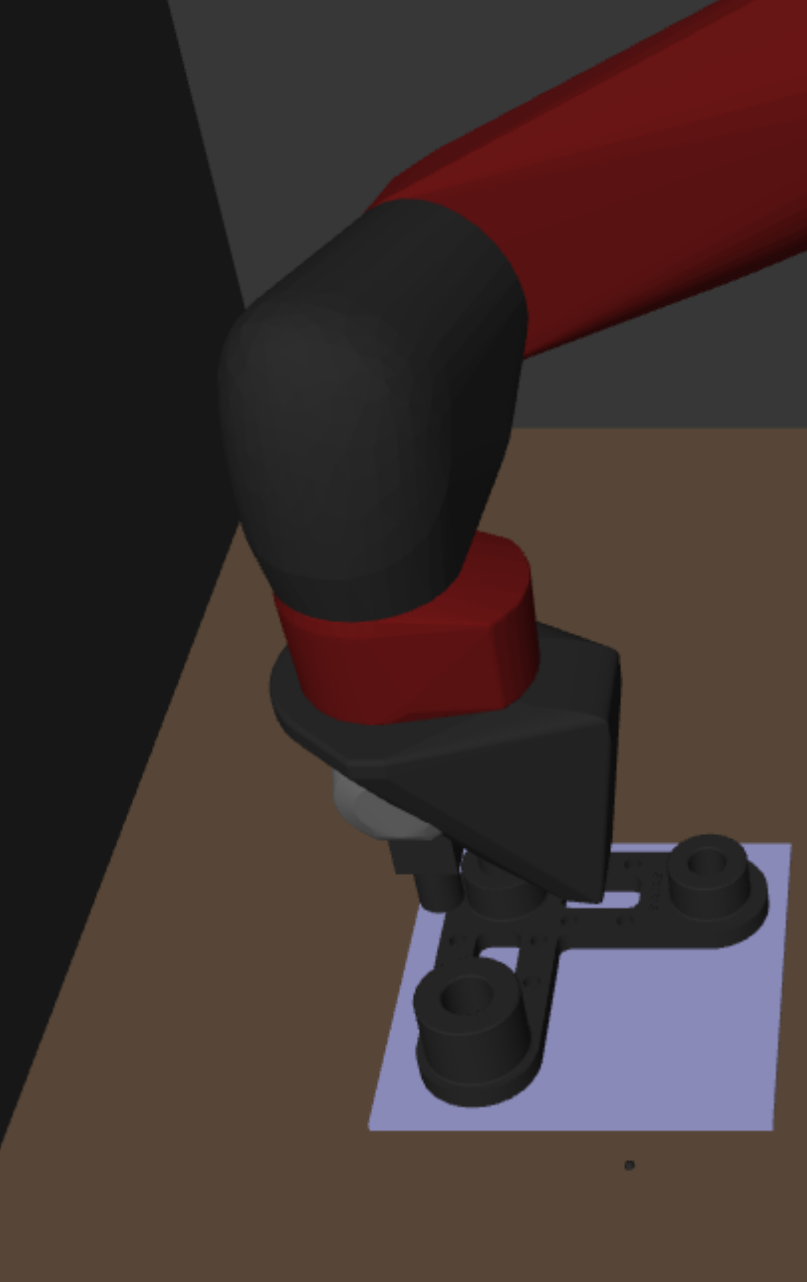}\includegraphics[width=0.515\textwidth]{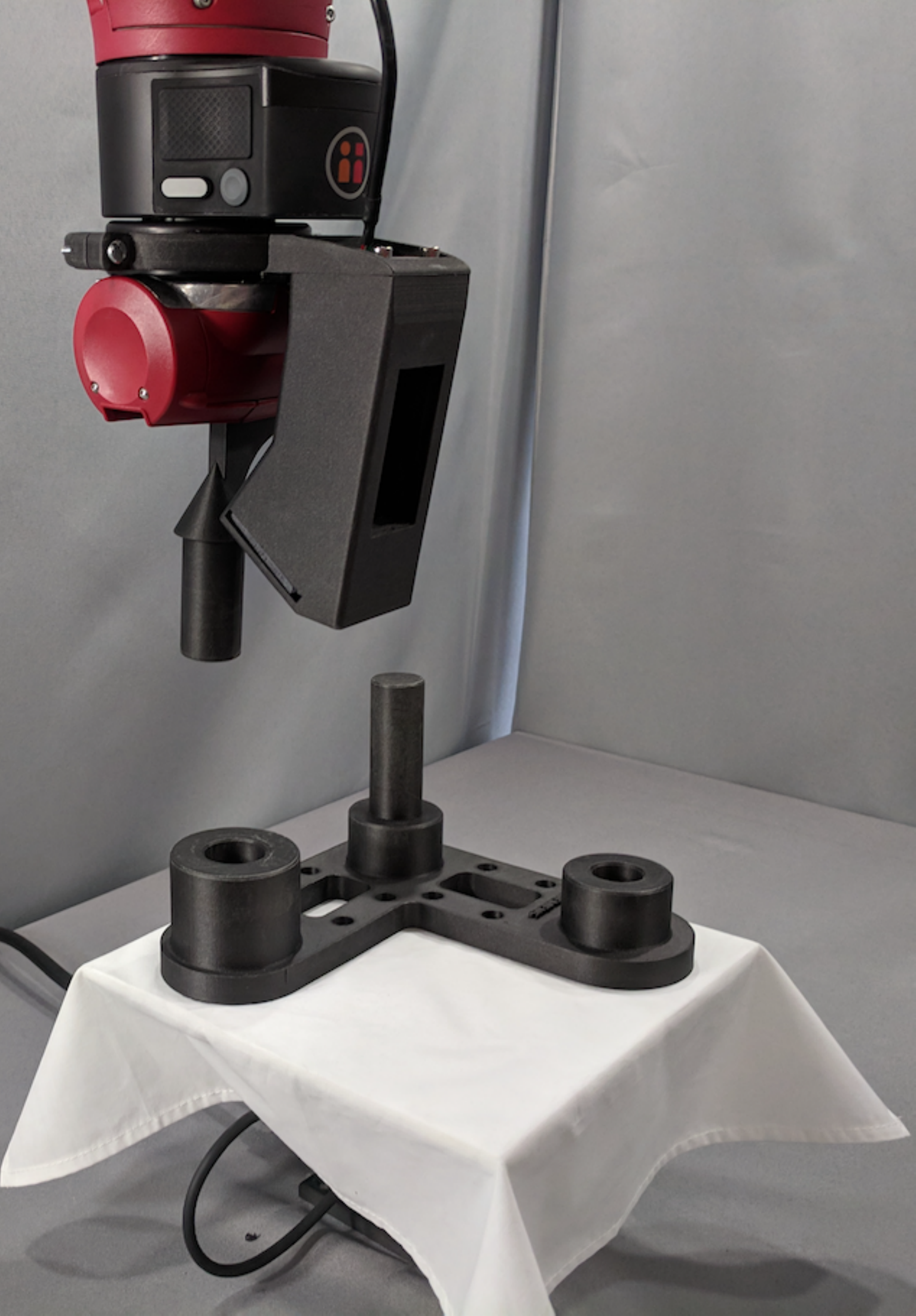}}
        \caption{}
        \label{fig:gears_robo}
    \end{subfigure}
    \begin{subfigure}[b]{0.33\textwidth}
        \center{\includegraphics[width=\textwidth, height=0.681\textwidth]{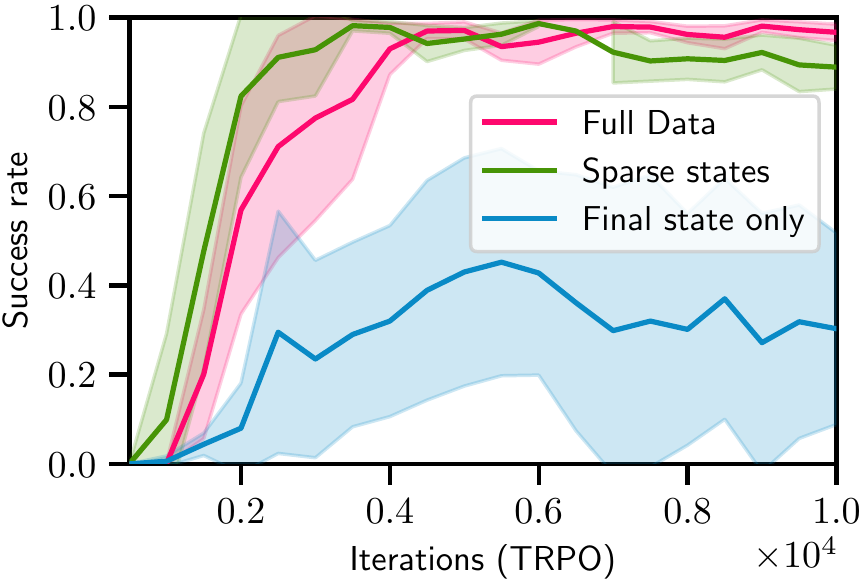}}%2.11787in
        \caption{}
        \label{fig:gears_gail}
    \end{subfigure}
    \begin{subfigure}[b]{0.33\textwidth}
        \center{\includegraphics[width=\textwidth, height=0.681\textwidth]{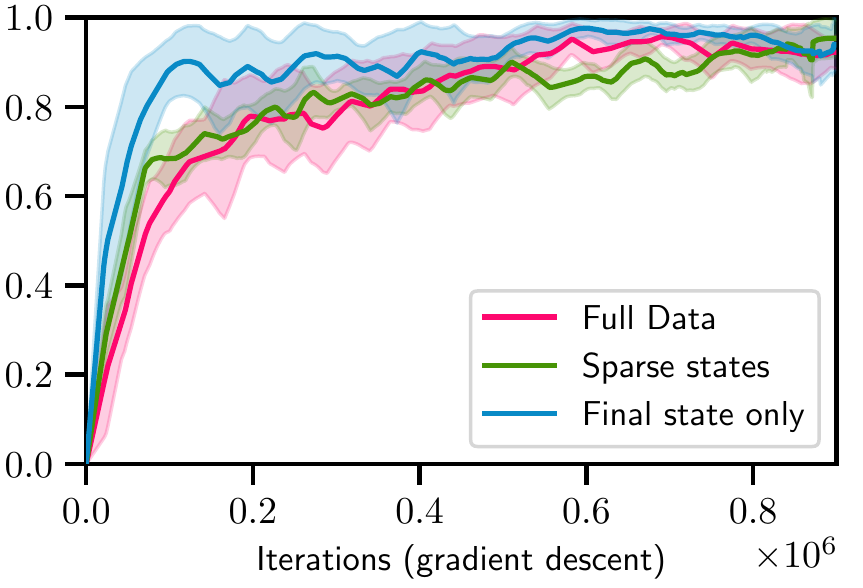}}
        \caption{}
        \label{fig:gears_gpril}
    \end{subfigure}
%\end{figure}
%\begin{figure}[t]
    \begin{subfigure}[b]{0.32\textwidth}
        \center{\includegraphics[width=\textwidth, height=0.681\textwidth]{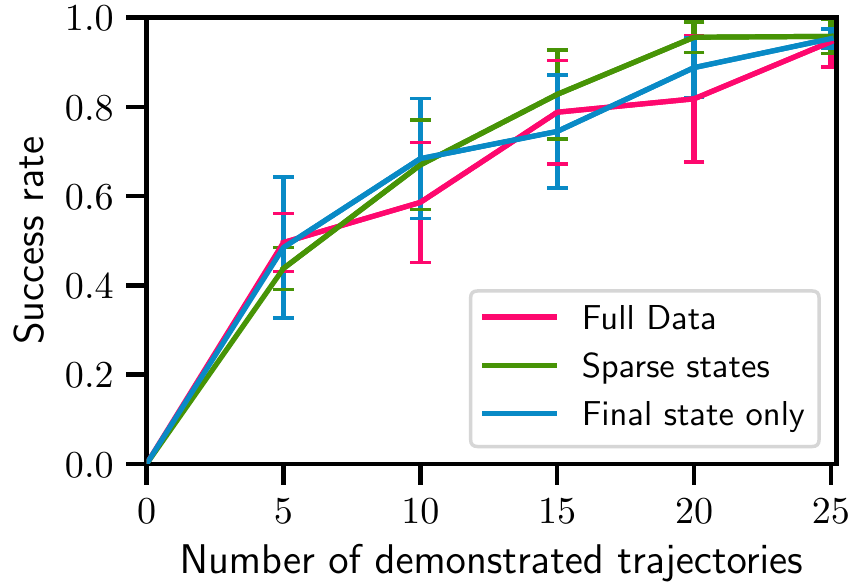}}\vspace*{-0.2cm}
        \caption{}\vspace*{-0.1cm}
        \label{fig:gears_nondemo}
    \end{subfigure}
    \begin{subfigure}[b]{0.32\textwidth}
        \center{\includegraphics[width=\textwidth, height=0.681\textwidth]{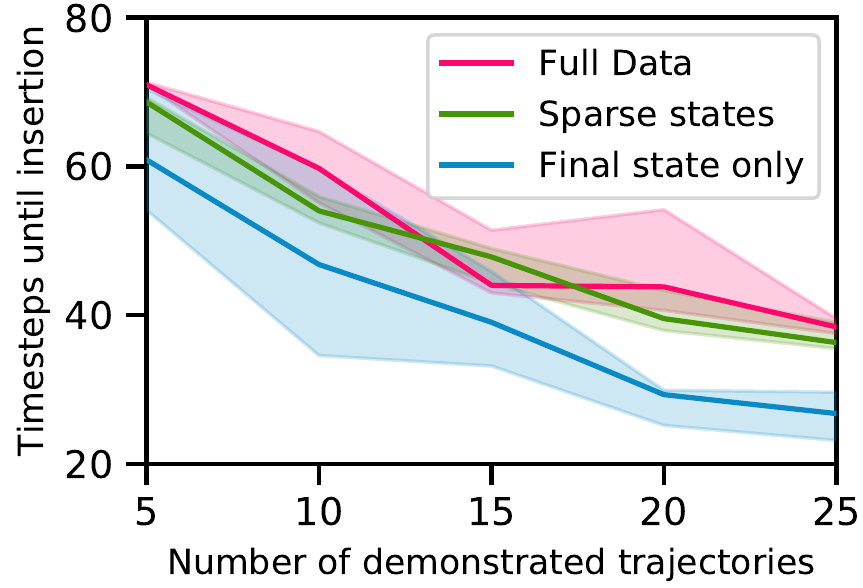}}\vspace*{-0.2cm}
        \caption{}\vspace*{-0.1cm}
        \label{fig:gears_efficiency}
    \end{subfigure}
    \begin{subfigure}[b]{0.32\textwidth}
        \center{\includegraphics[width=\textwidth, height=0.681\textwidth]{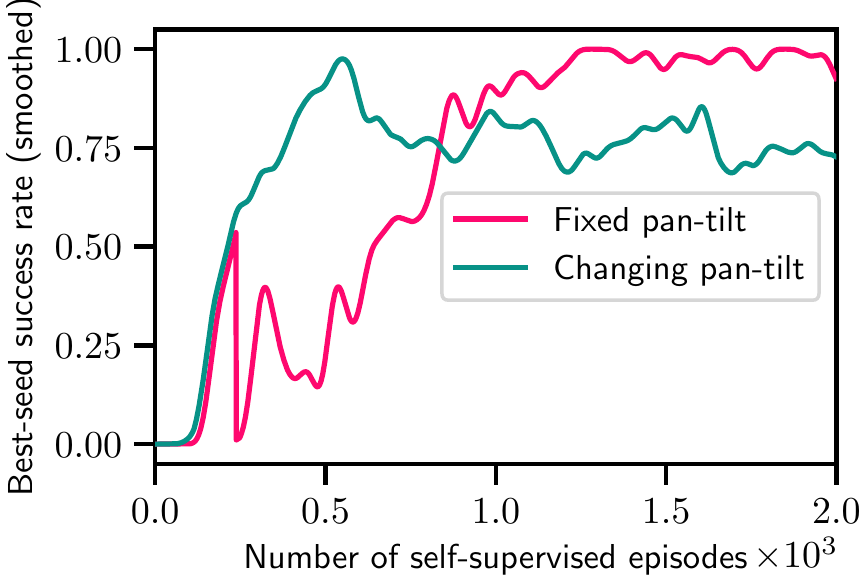}}\vspace*{-0.2cm}
        
        \caption{}\vspace*{-0.1cm}
        \label{fig:real_robo_curve}
    \end{subfigure}
    \caption{\textbf{a)} Depiction of the peg insertion task \textbf{b)} Average success rate and 95\% confidence interval of GAIL with 25 demonstrations across 10 runs (evaluated over 100 roll-outs). \textbf{c)} Average success rate of GPRIL across 5 seeds. Unlike GAIL, the performance of GPRIL doesn't drop off when provided with only final states.
\textbf{d)} Average success rate and confidence interval of GPRIL. Final performance after $10^6$  iterations increases steadily as the number of demonstrated trajectory increases but is unaffected by dropping steps from each demonstration. \textbf{e)} Median
length and IQR of trajectories that are successfully inserting the peg. Providing only final states is significantly
faster. \textbf{f)} Best seed performance on both variations of peg insertion on the real robot.}
%TODO: add analysis why dropping steps from demonstrations is better than dropping demonstrations
\end{figure}
\subsection{Clip insertion}\label{sec:eval_clippy}
% Task description
In the first task, a simulated robot arm has to insert an %is tasked to
elastic clip into a plug, which requires the robot to first flex the clip in order to be able to insert it (see figure \ref{fig:clippy_robo}). %TODO:term ??
In real-world insertion tasks, the pose of the robot, the socket, or the grasped object may vary.  We capture this
variability by mounting the socket on a pan-tilt unit, which is randomized by $\pm 0.8$ (pan) and $\pm 0.2$ radians
(tilt). % TODO: ask Jon| and utilize haptic data to ``feel'' for the opening.
To perform this behavior, the robot observes proprioceptive features, specifically joint position, velocity and torques as well as the position of the end-effector and the socket orientation as a unit quaternion. %TODO: correct?
The task terminates when the robot leaves the work-space, reaches the goal, or after 50 seconds.
% Network description and important parameters for both approaches. 

For comparative evaluation, we train a policy network to predict mean and variance, modelling a multivariate normal
distribution over target velocities and train it using GPRIL, GAIL as well as behavioral cloning. We record
expert demonstrations using tele-operation and normalize observations based on the recorded demonstrations. We then
train GPRIL using a single asynchronous simulation and compare against the open source implementation of
GAIL\footnote{https://github.com/openai/baselines/tree/master/baselines/gail} for which we use 16 parallel simulations.
We select the best hyper parameters for GAIL found on a grid around the
hyperparameters used by~\citet{ho2016b} but lower the batch size to 256 as it increases the learning speed and accounts for the significantly
slower simulation of the task. We furthermore enable bootstrapping regardless of whether or not the episode terminated.
As all discriminator rewards are positive, handling terminal transitions explicitly can induce a bias towards longer
episodes. This is beneficial in the domains used by Ho and Ermon but harmful
in domains such as ours where the task terminates on success. A detailed list of hyper-parameters can be found in
appendix \ref{sec:hypers}.

We report final results after convergence %5 million iterations of gradient descent for behavioral cloning, 2 million iterations for
%GPRIL and 10000 iterations of TRPO for GAIL in figure \ref{fig:clippy_numdemo}\jsnote{These numbers make GAIL sound way more efficient}. 
and can see in figure \ref{fig:clippy_numdemo} that both
GAIL and GPRIL outperform behavioral cloning, indicating that generalizing over state-action trajectories requires fewer
demonstrations than generalizing over actions alone. Furthermore, we observe a higher
success rate using GPRIL and find that policies trained using GPRIL are more likely to retry insertion if the robot
slides the clip past the insertion point. To compare sample efficiency of GPRIL to GAIL, we limit the rate at which the asynchronous actor is collecting data.
%\st{slow down} \jsnote{that sounds like an arbitrary handicap.  perhaps "both algorithms are asynchronous, so to compare their sample efficiency we coordinate the rate at which the actors collect data."?} the rate at which the actor collects new roll-outs. 
While sample efficiency of GAIL could be increased by decreasing batch size or increasing various learning rates, we found
that this can lead to unstable learning performance while reducing the amount of samples required by only a small amount.
As can be seen in figure \ref{fig:clippy_samples}, GPRIL requires several orders of magnitudes fewer environment
interactions to learn this task. Finally, we evaluate the case where the expert's actions are unknown. 
%This scenario can enable the expert to provide demonstrations using different input modalities such as kinesthetic teaching \cite{chernova_robot_2014}. 
Since the state-space includes information about joint velocities as well as positions,
we find that matching the state-distribution is sufficient to solve the task. GPRIL can achieve this by setting
$\beta_d=1$ and $\beta_\pi=0$ . As can be seen in figure \ref{fig:clippy_obs}, performance deteriorates only marginally
with a similar difference in performance between both methods.

\subsection{Peg insertion with partial demonstrations}\label{sec:eval_gears}
The second task is a simulated version of the peg-insertion task depicted in figure \ref{fig:gears_robo}.
In this task, the
robot has to insert the peg into the hole, which is again mounted on a pan-tilt that randomly assumes pan and tilt angles
varying by 0.4 and 0.1 respectively. %TODO angles.
Hyperparameters are largely identical and we report minor differences in appendix \ref{sec:hypers}. Observation and
action space are identical with the exception of the omission of torques from the observation space as they are not
necessary to solve this task. 
We use this task to evaluate the performance of GAIL and GPRIL when learning from only a very limited set of demonstrated
states. To this end, we compare three different scenarios in which the demonstrations are sparsified to varying degrees: In the first case, the agent has access to the full state-trajectories of the expert, in the second
only every tenth state is available and in the third the agent sees only the final state of each of the 25 trajectories. Being able to learn from
only partial demonstrations is a useful benchmark for the effectiveness of imitation learning methods but can also
provide a convenient way of providing demonstrations and can free the agent to find more optimal trajectories between
states (see for example~\citet{akgun_trajectories_2012, schroecker2016directing}). As can be seen in figures
\ref{fig:gears_nondemo} and \ref{fig:gears_efficiency}, GPRIL achieves
similar final success rates in all three scenarios while being able to learn a significantly faster insertion policy
when learning from final states alone. We find that in the first two scenarios, this holds for GAIL as well
as can been in figure \ref{fig:gears_gail} while in the third case, GAIL becomes highly unstable and the resulting
performance can vary wildly, leading to a low average success rate. We hypothesize that these instabilities are a result of the discriminator overfitting to the very small amount of negative samples in its training data.

\subsection{Peg insertion on a physical system}\label{sec:eval_robo}
In previous sections we demonstrated sample-efficiency that indicates applicability of GPRIL to real-world physical
systems. To test this, we evaluate our approach on two variations of the physical peg-insertion task depicted in figure
\ref{fig:gears_robo}, involving a physical pan-tilt unit which is fixed in one scenario and has pan and tilt angles varying by  $0.1$ and $0.02$ radians in the second scenario.
For each scenario, we provide 20 demonstrations using kinesthetic teaching, which constitutes a natural way of recording demonstrations but provides state-trajectories only~\citep{chernova_robot_2014}.
 %Solving this task requires that the algorithm can generalize across multiple approach angles, learn from states alone as well as be efficient in its use of recorded interactions with the environment.
 Hyper-parameters are altered from section \ref{sec:eval_gears} to trade off a small margin of accuracy for higher learning speeds and are reported in appendix \ref{sec:hypers}. Note, however, that tuning hyper-parameters precisely is very difficult on a physical system. As can be seen in figure \ref{fig:real_robo_curve}, GPRIL is able to learn a successful insertion policy that generalizes to unseen insertion angles using just a few hours of environment interactions\footnote{Total time to collect and train on 2000 roll-outs was 18.5 and 16.5 hours on the fixed and changing versions of the task respectively. However, GPRIL converged to good policies significantly sooner.}. We report best-seed performance as we observe a high amount of variability due to factors outside the agent's control, such as the pan-tilt unit not reporting accurate information after physical contact with the robot. However, we wish to point out that the increased difficulty due to less predictable control is also likely to introduce additional variance that could be reduced further with careful design of exploration noise and other hyper-parameters. We furthermore provide a video of the training procedure and final policy to highlight the efficiency of our method\footnote{https://youtu.be/Dm0OCNujEmE}. %We report best-seed performance to highlight the applicability of the algorithm but wish to point out the higher variability in final results due to a variety of factors, including factors controllable through hyper-parameters such as a sub-optimal form of exploration noise for this particular system as well as factors outside the agents control such as inaccurate reports of the pan-tilt units position after physical contact with the robot. 

\section{Conclusion}
We introduced GPRIL, a novel algorithm for imitation learning which uses generative models to model multi-step
predecessor distributions and to perform state-action distribution matching. We show that the algorithm compares favorably with
state-of-the-art imitation learning methods, achieving higher or equivalent performance while requiring several orders of magnitude fewer environment samples.
Importantly, stability and sample-efficiency of GPRIL are sufficient to enable experiments on a real robot, which we demonstrated on a peg-insertion task with a variable-position socket.
%We believe that modeling multi-step transitions provides a way to densely extract information from every
%environment interaction while avoiding the pitfalls of model-based learning and that the techniques developed in this work
%provide a principled way to utilize such models.

%\vfill{}
\pagebreak
\bibliography{library}
\bibliographystyle{humannat}

\appendix
\section{Derivation of equation \ref{eq:path_expectation}}\label{sec:derivation}
Here, we derive equation \ref{eq:path_expectation} which unrolls the recursive definition of $\nabla_\theta \log
d^{\pi_\theta}(\overline{s})$ and rewrites it such that it can be replaced by an expectation over states and actions
along trajectories leading to the state $\overline{s}$. 
In section \ref{sec:stationary_gradient}, we derive a recursive definition of  $\nabla_\theta \log
d^{\pi_\theta}(\overline{s})$ which we will restate in more detail:
\begin{equation*}\begin{aligned}
    \nabla_\theta d^{\pi_\theta}(\overline{s}) &= \int \nabla_\theta \rho^{\pi_\theta}(s)  p(s_{t+1}=\overline{s}|s_t=s, a_t=a)  ds,a\\
     d^{\pi_\theta}(\overline{s}) \nabla_\theta \log d^{\pi_\theta}(\overline{s}) &= \int\rho^{\pi_\theta}(s, a)  p(s_{t+1}=\overline{s}|s_t=s, a_t=a)  \nabla_\theta \log  \rho^{\pi_\theta}(s, a) ds,a\\
   % p(s_{t+1}=\overline{s}|s_t=s, a_t=a)\left(\nabla_\theta \log d^{\pi_\theta}(s) + \nabla_\theta \log \pi_\theta(a|s)
  %  \right) ds,a%\\
    \nabla_\theta \log d^{\pi_\theta}(\overline{s}) &= \int q^{\pi_\theta}(s_{t}=s, a_t=a|s_{t+1}=\overline{s})\left(\nabla_\theta \log d^{\pi_\theta}(s) + \nabla_\theta \log \pi_\theta(a|s) \right) ds,a
\end{aligned}\end{equation*}
We can now unroll this definition:
\begin{equation}
    \begin{aligned}
        \nabla_\theta \log d^{\pi_\theta}(\overline{s}) =&  \mathrlap{\int q^{\pi_\theta}(s_{t}=s,a_t=a|s_{t+1}=\overline{s})\left(\nabla_\theta \log d^{\pi_\theta}(s) + \nabla_\theta \log \pi_\theta(a|s) \right) ds,a}\\
                                                        =& \lim_{T \rightarrow \infty} &&\mkern-18mu\int &&\mkern-18mu\left(q^{\pi_\theta}(s_{t+T-1}, a_{t+T-1}|s_{t+T} = \overline{s}) \prod_{j=0}^{T-2} q^{\pi_\theta}(s_{t+j}, a_{t+j}|s_{t+j+1}) \vphantom{\sum_{j=0}^{T} \nabla_\theta \log \pi_\theta(a_{t+j}|s_{t+j}) } \right.
                                                       \\&                             &&\mkern-18mu     &&\mkern-18mu\left.\vphantom{ q^{\pi_\theta}(s_{t+T-1}, a_{t+T-1}|s_{t+T} = \overline{s}) \prod_{j=0}^{T-2} q^{\pi_\theta}(s_{t+j}, a_{t+j}|s_{t+j+1})}\sum_{j=0}^{T} \nabla_\theta \log \pi_\theta(a_{t+j}|s_{t+j}) \right)ds_{t:t+T-1},a_{t:t+T-1}
                                                     \;+ \\&                             &&\mkern-18mu\mathrlap{\int q^{\pi_\theta} (s_t=s, a_t=a|s_{t+T} = \overline{s}) \nabla_\theta \log d^{\pi_\theta} (s)  ds, a}
    \end{aligned}
\end{equation}
Note that $\lim_{T \rightarrow \infty} q^{\pi_\theta}(s_t=s, a_t=a| s_{t+T}=\overline{s}) = \rho^{\pi_\theta}(s, a)$ due to Markov
chain mixing and, therefore, the second term of the above sum reduces to 0 as \begin{equation}\int d^{\pi_\theta}(s)
\pi(a|s)\nabla_\theta \log d^{\pi_\theta}(s) ds, a = 0.\end{equation}
    By pulling out the sum, we can now marginalize out most variables and shift indices to arrive at the desired
    conclusion:
\begin{equation}\label{eq:path_expectation_deriv}
    \begin{aligned}
        \nabla_\theta \log d^{\pi_\theta}(\overline{s}) =& \lim_{T \rightarrow \infty} 
        \sum_{j=0}^T \int q^{\pi_\theta}(s_{t + j}=s, a_{t+j}=a| s_{t+T+1} = \overline{s}) \nabla_\theta \log
        \pi_\theta(a|s) ds,a\\
        =&  \lim_{T \rightarrow \infty} 
        \sum_{j=0}^T \int q^{\pi_\theta}(s_{t}=s, a_{t}=a| s_{t+T+1-j} = \overline{s}) \nabla_\theta \log
        \pi_\theta(a|s) ds,a\\
        =& \lim_{T \rightarrow \infty} 
        \int \sum_{j=0}^T  q^{\pi_\theta}(s_{t}=s, a_{t}=a| s_{t+j+1} = \overline{s}) \nabla_\theta \log
        \pi_\theta(a|s) ds,a
    \end{aligned}
\end{equation}
\vfill\pagebreak
\section{Hyperparameters}\label{sec:hypers}
\begin{table}[h]
    \begin{subtable}{0.49\textwidth}
    \centering \small
        \begin{tabular}{|l|l|}
            \hline
            \multicolumn{2}{|l|}{\bf General parameters} \\\hline
            Total iterations & $2e6$ \\\hline 
            Batch size $B$ & 256\\\hline
            $\gamma$ & 0.9 \\\hline
            Replay memory size & 50000 \\\hline
            $N_{\mathcal{B}}$ & 15000 \\\hline
            $N_{\pi}$ & 5000 \\\hline
            \multicolumn{2}{|l|}{\bf $\mathcal{B}^s$ and $\mathcal{B}^a$} \\\hline
            Stacked autoencoders & 2\\\hline
            Hidden layers & 500, 500\\\hline
            Optimizer & Adam\\\hline
            Learning rate & $2\cdot 10^5$\\\hline
            Burnin & 50000 iterations \\\hline
            L2-regularization & $10^{-2}$\\\hline
            $min(\sigma_i)$ & 0.1\\\hline %TODO: mention cmaf normalizers?
            Gradient clip, $L_2$ norm & 100\\\hline
            \multicolumn{2}{|l|}{\bf Policy $\pi_\theta$} \\\hline
            Hidden layers & 300, 200 \\\hline
            Optimizer & Adam\\\hline
            Learning rate & $10^4$\\\hline
            $\sigma$ bounds & (0.01, 0.1) \\\hline %TODO: mention action output normalizers
        \end{tabular}
        \caption{GPRIL parameters for clip insertion}
    \end{subtable}
    \begin{subtable}{0.49\textwidth}
        \centering \small
        \begin{tabular}{|l|l|}
            \hline
            \multicolumn{2}{|l|}{\bf General parameters} \\\hline
            Total iterations & $1e4$ \\\hline 
            \#Processes & 16\\\hline
            Batch size $B$ & $16\cdot256$\\\hline
            Actor steps per iteration & 3\\\hline
            Discriminator steps per iteration & 1\\\hline
            $\gamma$ & 0.995\\\hline
            \multicolumn{2}{|l|}{\bf Actor} \\\hline
                Hidden layers & 300, 200 \\\hline
                KL step size & 0.01\\
                $\sigma$ bounds & (0.01, 0.1) \\\hline
            \multicolumn{2}{|l|}{\bf Discriminator} \\\hline
                Hidden layers & 150, 100 \\\hline
                Optimizer & Adam\\\hline
                Learning rate & $10^4$\\\hline
                Entropy regularization & 1 \\\hline
            Optimizer & Adam\\\hline
            \multicolumn{2}{|l|}{\bf Critic} \\\hline
                Hidden layers & 300, 200 \\\hline
                Optimizer & Adam\\\hline
                Learning rate & $5\cdot10^3$\\\hline
        \end{tabular}
        \caption{GAIL parameters for clip insertion}
    \end{subtable}
    \begin{subtable}{\textwidth}
        \centering \small \vspace{0.3cm}
        \begin{tabular}{|l|l|}
            \hline
            \multicolumn{2}{|l|}{\bf General parameters} \\\hline
            Total iterations & $5e6$ \\\hline 
            Batch size $B$ & 256\\\hline
            Hidden layers & 300, 200 \\\hline
            $\sigma$ bounds & (0.01, 0.1) \\\hline
            Optimizer & Adam\\\hline
            Learning rate & $10^4$\\\hline
            L2-regularization & $10^{-4}$\\\hline
        \end{tabular}
        \caption{BC parameters for clip insertion} \vspace{0.3cm}
    \end{subtable}
    \begin{subtable}{0.55\textwidth}
        \centering  \small
        \begin{tabular}{|l|l|l|}
            \hline
            & {\bf Simulation} & {\bf Real robot} \\\hline
            \multicolumn{3}{|l|}{\bf General parameters} \\\hline
            Batch size $B$ & \multicolumn{2}{|l|}{256}\\\hline
            $\gamma$ & 0.9 & 0.7 \\\hline
            Replay memory size & 10000 & 50000 \\\hline
            $N_{\mathcal{B}}$ & \multicolumn{2}{|l|}{10000} \\\hline
            $N_{\pi}$ & 1000 & 5000 \\\hline
            \multicolumn{3}{|l|}{\bf $\mathcal{B}^s$ and $\mathcal{B}^a$} \\\hline
            Stacked autoencoders & \multicolumn{2}{|l|}{2}\\\hline
            Hidden layers & \multicolumn{2}{|l|}{500, 500}\\\hline
            Optimizer & \multicolumn{2}{|l|}{Adam}\\\hline
            Learning rate & $10^{-5}$ & $3\cdot 10^{-5}$\\\hline
            Burnin & \multicolumn{2}{|l|}{0} \\\hline
            L2-regularization & $10^{-2}$ & $10^{-3}$\\\hline
            $min(\sigma_i)$ & 0.1 & 0.01\\\hline %TODO: mention cmaf normalizers?
            Gradient clip ($L_2$) & \multicolumn{2}{|l|}{100}\\\hline
            \multicolumn{3}{|l|}{\bf Policy $\pi_\theta$} \\\hline
            Hidden layers & \multicolumn{2}{|l|}{300, 200} \\\hline
            Optimizer & \multicolumn{2}{|l|}{Adam}\\\hline
            Learning rate & \multicolumn{2}{|l|}{$10^{-4}$}\\\hline
            $\sigma$ bounds & \multicolumn{2}{|l|}{(0.01, 0.1)} \\\hline %TODO: mention action output normalizers
        \end{tabular}
        \caption{GPRIL parameters for peg insertion}
    \end{subtable}
    \begin{subtable}{0.44\textwidth}
        \centering  \small
        \begin{tabular}{|l|l|}
            \hline
            \multicolumn{2}{|l|}{\bf General parameters} \\\hline
            \#Processes & 16\\\hline
            Batch size $B$ & $16\cdot256$\\\hline
            Actor steps/iteration & 3\\\hline
            Discriminator steps/iteration & 1\\\hline
            $\gamma$ & 0.995\\\hline
            \multicolumn{2}{|l|}{\bf Actor} \\\hline
                Hidden layers & 300, 200 \\\hline
                KL step size & 0.01\\
                $\sigma$ bounds & (0.01, 0.1) \\\hline
            \multicolumn{2}{|l|}{\bf Discriminator} \\\hline
                Hidden layers & 150, 100 \\\hline
                Optimizer & Adam\\\hline
                Learning rate & $10^{-4}$\\\hline
                Entropy regularization & 1 \\\hline
            Optimizer & Adam\\\hline
            \multicolumn{2}{|l|}{\bf Critic} \\\hline
                Hidden layers & 300, 200 \\\hline
                Optimizer & Adam\\\hline
                Learning rate & $5\cdot10^{-3}$\\\hline
        \end{tabular}
        \caption{GAIL parameters for simulated peg insertion}
    \end{subtable}
\end{table}
\section{Relation to the Policy Gradient Theorem}\label{sec:appendix_discount}
In this section, we outline the relation between the policy gradient theorem~\citep{sutton1999policy} and the state-action-distribution gradient derived in section \ref{sec:derivation} and show equivalence of the discount factor used in reinforcement learning and the discount factor $\gamma$ introduced in this work. We first show that the state-action distribution gradient is equal to the policy gradient in the average reward case using a specific reward function. We then derive the $\gamma$-discounted approximation of the state-action-distribution gradient presented in section \ref{sec:stationary_gradient} from the policy gradient in the discounted reward framework using $\gamma$ as a discount factor. While this derivation is more cumbersome than the one presented in the body of the paper, it allows us to gain a better understanding of the meaning of $\gamma$ since discounting in the reinforcement learning setting is well understood.
For notational simplicity, we assume that the state space $\mathcal{S}$ and action space $\mathcal{A}$ are countable. Note that this is a common assumption made in the field~\citep{tsitsiklis1997analysis} and considering that any state-action space is countable and finite when states and actions are represented using a finite number of bits makes it apparent that this assumption does not constitute a simplification in the practical sense.

In this work we propose to follow the gradient $\nabla_\theta \log \rho^{\pi_\theta}$ evaluated at demonstrated states and actions, consider thus the gradient evaluated at $\overline{s}, \overline{a}$:
\begin{equation}\begin{aligned}
\nabla_\theta \log \rho^{\pi_\theta} (\overline{s}, \overline{a}) &= \frac{1}{\rho^\pi_\theta(\overline{s}, \overline{a})} \nabla_\theta \sum_{s\in \mathcal{S}, a \in \mathcal{A}}  \rho^{\pi_\theta} (s, a) \mathbf{1}(s=\overline{s}, a=\overline{a})\\
&=\nabla_\theta \mathbb{E}_{s,a\sim \rho^\pi_\theta}\left[R(s,a)\right]=:\nabla_\theta J(\pi_\theta)
\end{aligned}
\end{equation}
As we can see, the gradient is equivalent to the policy gradient using the reward function ${R(s,a) :=
    \frac{\mathbf{1}(s=\overline{s}, a=\overline{a})}{\rho^\pi_{\hat{\theta}}(\overline{s}, \overline{a})}}$ in the average reward framework where $\hat{\theta}$ corresponds to the parameters of the policy at the current iteration. This reward function is not practical as it can be infinitely sparse and furthermore depends on the unknown stationary distribution. However, it allows us to derive the policy gradient using this reward function in the discounted reward setting which constitutes a well understood approximation of the average-reward scenario. Following the notation in~\citet{sutton1999policy}, we have:
\begin{equation}
\begin{aligned}
\nabla_\theta J(\pi_\theta) &=  \mathbb{E}_{s_j,a_j\sim \rho^\pi_\theta}\left[\nabla_\theta \log \pi_\theta(a_j|s_j) \mathbb{E}_{s_{j+1}, a_{j+1}, \dots}\left[ \sum_{t=0}^\infty \gamma^t R(s_{t+j},a_{t+j})|\pi_\theta, s_j, a_j\right]\right]
\end{aligned}
\end{equation}
We can now replace the expectation over the stationary distribution by an expectation over the path of the agent. This yields a double sum whose order of summation can be changed:
\begin{equation}
\begin{aligned}
\nabla_\theta J(\pi_\theta) &= \lim_{T\rightarrow \infty} \frac{1}{T} \mathbb{E}\left[\sum_{j=0}^T \nabla_\theta \log \pi_\theta(a_j|s_j) \sum_{t=0}^T \gamma^t R(s_{t+j},a_{t+j})|\pi_\theta \right]\\
&= \lim_{T\rightarrow \infty} \frac{1}{T} \mathbb{E}\left[\sum_{j=0}^T \sum_{t=j}^T \nabla_\theta \log \pi_\theta(a_j|s_j) \gamma^{t-j} R(s_{t},a_{t})|\pi_\theta \right]\\
&= \lim_{T\rightarrow \infty} \frac{1}{T} \mathbb{E}\left[\sum_{t=0}^T\sum_{j=0}^t  \gamma^{t-j} \nabla_\theta \log \pi_\theta(a_j|s_j) R(s_{t},a_{t})|\pi_\theta \right]
\end{aligned}
\end{equation}
After changing the order of summation we can replace the outer sum with the expectation over the stationary distributions and use the special nature of the chosen reward function to write the gradient as a conditional expectation:
\begin{equation}
\begin{aligned}
\nabla_\theta J(\pi_\theta)&=  \lim_{t\rightarrow \infty} \mathbb{E}\left[\sum_{j=0}^t \gamma^{t-j} \nabla_\theta \log \pi_\theta(a_j|s_j)  R(s_{t},a_{t})|\pi_\theta \right]\\
&= \lim_{t\rightarrow \infty} \mathbb{E}\left[\sum_{j=0}^t \gamma^{t-j} \nabla_\theta \log \pi_\theta(a_j|s_j)|\pi_\theta, s_t=\overline{s}, a_t=\overline{a} \right]
\end{aligned}
\end{equation}
Finally, we notice that this equation constitutes the discounted version of equation \ref{eq:path_expectation_deriv}, thus we can immediately obtain our estimate of the state-action-distribution gradient:
\begin{equation}
\begin{aligned}
\nabla_\theta J(\pi_\theta) &= \frac{1}{1-\gamma} \mathbb{E}_{s, a\sim \mathcal{B}^{\pi_\theta}(\cdot, \cdot|\overline{s})}\left[\nabla_\theta \log \pi_\theta(a|s)\right] + \nabla_\theta \log \pi_\theta(\overline{a}|\overline{s})\approx \nabla_\theta \log \rho^{\pi_\theta} (\overline{s}, \overline{a})
\end{aligned}
\end{equation}
While this derivation is less direct than the derivation used in section \ref{sec:derivation}, it draws a connection between the problem of matching state-action-distributions and the reinforcement learning problem with a reward that is positive for demonstrated state-action pairs and 0 otherwise. This indicates that the role of the discount factor is similar in both settings: Lower values of $\gamma$ trade-off accurate matching of distributions for lower variance and as expedience. Agents with low $\gamma$ will learn policies that recover and reach demonstrated states quicker over policies that are matching the experts state-action distribution more accurately long-term. Furthermore, the long history of successes in reinforcement learning validates the use of $\gamma$ as an approximation to the true objective which is often more accurately described by the average-reward objective. 
\end{document}

%% file: math_commands.tex
\usepackage{amsmath,amsfonts,bm}

\def\eqref#1{equation~\ref{#1}}
\def\1{\bm{1}}

\DeclareMathAlphabet{\mathsfit}{\encodingdefault}{\sfdefault}{m}{sl}
\SetMathAlphabet{\mathsfit}{bold}{\encodingdefault}{\sfdefault}{bx}{n}